# Smart Operation Theatre

Saraf Krish
School of Social Science

Dr Cai Yiyu
Dr Huang Li Hui
School of Mechanical and Aerospace Engineering

***Abstract -*** During surgeries, there's a risk of medical gauzes being left inside patients' bodies, leading to "Gossypiboma" in patients and can cause serious complications in patients and also lead to legal problems for hospitals from malpractice lawsuits and regulatory penalties Diagnosis depends on imaging methods such as X-rays or CT scans, and the usual treatment involves surgical excision. Prevention methods, such as manual counts and RFID-integrated gauzes, aim to minimize gossypiboma risks. However, manual tallying of 100s of gauzes by nurses is time-consuming and diverts resources from patient care.

In partnership with Singapore General Hospital (SGH) we have developed a new prevention method, an AI-based system for gauze counting in surgical settings. Utilizing real-time video surveillance and object recognition technology powered by YOLOv5, a Deep Learning model was designed to monitor gauzes on two designated trays labelled "In" and "Out". Gauzes are tracked from the "In" tray, prior to their use in the patient's body & in the "Out" tray post-use, ensuring accurate counting and verifying that no gauze remains inside the patient at the end of the surgery. We have trained it using numerous images from Operation Theatres & augmented it to satisfy all possible scenarios.

This study has also addressed the shortcomings of previous project iterations. Previously, the project employed two models: one for human detection and another for gauze detection, trained on a total of 2800 images. Now we have an integrated model capable of identifying both humans and gauzes, using a training set of 11,000 images. This has led to improvements in accuracy and increased the frame rate from 8 FPS to 15 FPS now. Incorporating doctor's feedback, the system now also supports manual count adjustments, enhancing its reliability in actual surgeries.

**Keywords –**Gossypiboma, Retained surgical items (RSI), AI-based system for gauze counting, YOLOv5, Deep Learning model, RFID-integrated gauzes.

## 1 INTRODUCTION

When performing surgeries, doctors need to use medical gauze to stop bleeding at surgery sites. Sometimes, doctors forget to remove the gauze from the patient's body, causing it to stay inside and potentially lead to pain, abscess formation, septic shock or even death. This condition, Gossypiboma is where cotton-based surgical sponges are accidentally left inside a patient's body after surgery. Treating this condition typically requires surgically removing the retained sponge and addressing any related complications. Additionally, among the current prevention methods are manual counting, RFID-enabled gauzes, and X-ray detectable gauzes. RFID (Radio Frequency Identification) is a technology used for tracking objects and while RFID and X-ray detectable gauzes can significantly reduce human errors, they are significantly more expensive than standard gauzes. As a result, the majority of hospitals continue to rely on manual counting, with errors being inevitable.

This study assumes that AI technology, specifically object detection techniques, can be trained to accurately identify and track gauze during surgical procedures. Consequently, this research aims to investigate a novel method that leverages AI, deep learning, and computer vision technologies to improve the precision and cost-effectiveness of gauze tracking in surgical environments.

### 1.1 SCOPE & COLLABORATION

To guarantee precise outcomes and mitigate any potential risks to patient safety, this research will be carried out in partnership with Singapore General Hospital (SGH) for data gathering and conclusive trials, with oversight provided by doctors and nurses and– we will use sterilized and blood-stained gauzes from operations to train our model. Medical professionals have delineated various scenarios that may arise during surgical procedures. To address these, we have documented our image collection within actual operational theatres, capturing a comprehensive array of situations, including overlapping gauzes, shadows, and excessive lighting, to replicate the dynamics of live operations. Currently, we are conducting trials with our finalized model in actual surgeries to identify and address potential limitations, aiming to refine and enhance its performance.

An expected annual cost of $6.72 million is incurred by SingHealth Group due to gauze retained in the human body after surgery. The primary goal is to develop a highly accurate product for hospital deployment that detects retained gauzes. This innovation will enable hospitals to move away from manual counting, thereby reducing errors, improving patient safety, and lowering hospital costs.

## 1.2 THEORY AND LIMITATION

The study is grounded in AI and deep learning theories acting upon real-time object detection models like YOLOv5 and getting results from it. It also considers cost-efficiency theories related to healthcare technology adoption.

However, the study has limitations. There is potential variability in the AI model's accuracy due to differing operating conditions in different hospitals. Additionally, the study is limited to gauze tracking and does not address other types of retained surgical items. There is also potential to improve with newer models like yolov9 to enhance the detection speed.

## 2 LITERATURE REVIEW

Recent studies on automated gauze counting systems with RFID-integrated gauzes have shown a 100% detection accuracy, confirming their reliability for gauze detection. Nevertheless, the widespread implementation of this system is impeded by cost concerns. The cost of a standard surgical gauze is approximately USD $0.07 per unit. Introducing an RFID system entails extra expenses, such as a scanner costing around $400.00 and an additional $0.50 for each RFID-tagged gauze.[1] This results in an RFID-embedded gauze being roughly eight times more expensive than a regular one, prompting most hospitals to continue using manual gauze counting to avoid the increased costs.

In a study examining 61 surgical cases, randomly selected from medical records with claims or reports of retained surgical items (RSI), where manual gauze counts were performed, it was found that 88 percent of these cases had a final count that was inaccurate yet considered correct, underscoring the significant risk of human error in manual counting [2]. Recently, techniques in computer vision, particularly those utilizing deep learning, have become prominent for identifying and counting gauze materials during surgeries. Numerous studies have explored the automatic detection of surgical gauzes through computer vision.

In June 2015, Álvaro García-Martínez and his team introduced a computer vision method for detecting surgical gauzes left inside patients post-surgery. Their method employs a texture-based algorithm derived from Local Binary Pattern (LBP) to identify gauze areas in images from laparoscopic cameras. This approach achieved a sensitivity of 42.95% and a specificity of 90.88%. [3]

Eusebio de la Fuente López and his team, in their paper 'Automatic gauze tracking in laparoscopic surgery using image texture analysis,' experimented with both LBP and Convolutional Neural Network (CNN) approaches. The LBP algorithm demonstrated robust gauze detection with 98% precision and 94% sensitivity, while the CNN approach achieved 100% precision and 97% sensitivity, though real-time processing was not feasible with standard hardware. [4]

More recently, in 2022, Guillermo Sánchez-Brizuela and his team used 4003 hand-labeled images from laparoscopic videos containing surgical gauze to train various CNN models. Their findings indicated that a U-NET-BASED architecture performed best for gauze segmentation, achieving an intersection over union of 0.85 and operating at over 30 fps. They also tested Yolov3, which could run in real-time but with moderate recall. [5]

Overall, the detection and counting of gauze using computer vision techniques have garnered significant interest. Deep learning methods, especially those based on CNNs, have outperformed traditional techniques like the LBP algorithm. Apart from these studies, NTU has previously conducted research utilizing YOLO architecture to achieve real-time object detection. This study improves upon the model created in Ding Jishen's paper, which focused on creating a gauze counting system. [6]

## 3 METHODOLOGY

The dataset used to train the AI-driven gauze tracking system was meticulously curated in collaboration with Singapore General Hospital (SGH). Using a dual-tray setup labeled "In" for unused gauze and "Out" for gauze stained with blood, images were captured from video footage recorded by cameras installed on the system within a real operating room. Under the guidance of a doctor, gauzes were handled under different lighting conditions and placed in various positions on the trays to simulate real surgical situations. These images were then annotated using Roboflow, a comprehensive platform for computer vision datasets, which facilitated efficient labelling of gauzes and hands. The annotated dataset was further augmented to increase its size and diversity, enhancing the model's ability to generalize across different scenarios. This extensive dataset was used to train a YOLOv5-based neural network, ensuring precise and reliable detection and tracking of gauzes in real-time during surgical procedures. The ultimate Python application utilizes this AI model to meticulously monitor the movement of gauze through accurate object detection and live tracking, providing continuous counts for both trays and greatly enhancing surgical safety and efficiency.

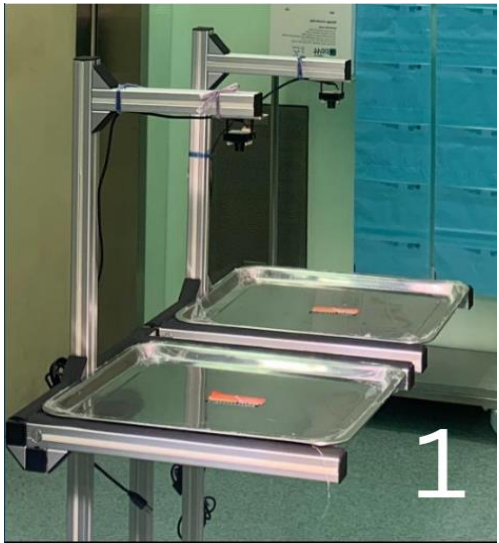
Figure 1. Dual Tray System

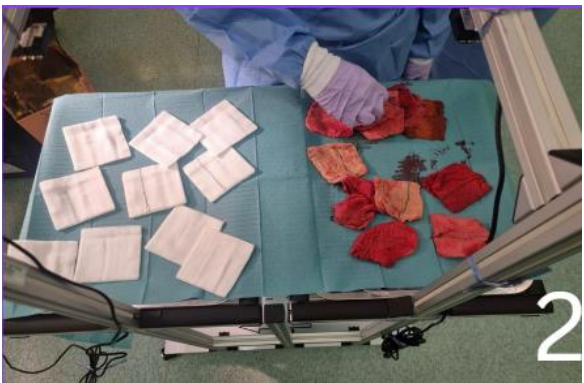
Figure 2. Collecting Gauze Images

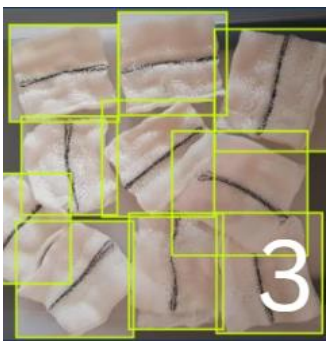
Figure 3. Annotation of gauzes

Figure 4. Training yolo model

Figure 5. Algorithm to Count gauzes

### 3.1 HARDWARE

The hardware of the AI-based surgical gauze management system consists of the following components:

1. Hardware Frame: The hardware frame of the AI-based surgical gauze management system features a structure designed to be stable and durable, typically made from lightweight yet sturdy materials like aluminum profiles. It includes adjustable arms that hold the cameras, allowing them to be positioned at optimal angles for capturing images of the gauzes on the trays. Additionally, the frame is equipped with a secure mounting mechanism, enabling it to be easily positioned over standard surgical trays ensuring minimal space usage and seamless integration into existing surgical setups. The frame is also equipped with wheels allowing it to be easily moved and positioned as needed within the operating theatre, enhancing the system's practicality and mobility within the OT.

2. Camera and Lighting: The system includes high-resolution cameras on adjustable arms to capture real-time video feeds of the gauzes on the trays, essential for accurate AI detection and counting. Integrated LED lights ensure consistent, clear illumination for high-quality image capture, regardless of ambient lighting conditions in the operating theatre.

3. Trays: The system uses standard surgical trays used to hold the gauzes to be counted and are designed to be easily replaceable and sterilizable

### 3.2 TECHNOLOGIES USED

This study leverages various advanced technologies to address the challenge of accurately tracking surgical gauze in operating theatres. Our code integrates several critical technologies, each playing a specific role in the development and deployment of the gauze tracking system. Below is a detailed explanation of these technologies and their relevance to the project.

**PyTorch**: It is a publicly accessible ML framework designed for tasks like computer vision and natural language processing. This library offers a versatile and effective platform for creating deep learning models.

Use Case: In this project, PyTorch is used to load and run the YOLOv5 model, which is crucial for the real-time detection and classification of objects (surgical gauzes and hands) in the video feeds from the operating theatre.

**YOLOv5 (You Only Look Once):** It is an advanced object detection model renowned for its swift processing and precision. By analyzing images in one go, it is ideal for applications requiring real-time performance.

Use Case: YOLOv5's ability to detect objects quickly and accurately is essential for the real-time monitoring of surgical gauzes. The model is trained to recognize and differentiate between gauzes and hands, ensuring precise tracking and counting during surgical procedures.

**Computer Vision (OpenCV):** It is a library of programming functions widely used for image and video processing in real time.

Use Case: OpenCV is used in this project to handle video capture, frame resizing, and create the user interface for output display operations. It allows the system to process and visualize video streams from the cameras, overlay bounding boxes, and text annotations for detected objects, and handle user inputs for interactive functionalities.

**CUDA (Compute Unified Device Architecture):** It is a platform for parallel computing and a programming model created by NVIDIA to facilitate general-purpose computing on GPUs(Graphics Processing Units), enabling accelerated computing applications.

Use Case: CUDA is utilized to accelerate the deep learning model's inference process, enabling real-time detection and tracking of gauzes in high-resolution video streams. This ensures the system can keep up with the demands of a busy operating theatre.

**Multithreading**: it is a programming technique that allows concurrent execution of two or more threads for maximizing the utilization of CPU resources. It helps in performing multiple operations simultaneously.

Use Case: In this project, multithreading is employed to manage video capture and processing from two cameras simultaneously. This ensures that both video streams are processed in real-time without causing delays, crucial for accurate and timely gauze tracking and improving FPS(Frames per Second)

Each of these technologies plays a critical role in the successful implementation of the AI-driven gauze tracking system. The integration of PyTorch and YOLOv5 provides the backbone for high-accuracy object detection. OpenCV handles the real-time video processing and user interaction, while CUDA ensures that the system can perform these tasks at the necessary speed. Multithreading allows for efficient simultaneous processing of multiple video streams, which is crucial in a dynamic operating theatre environment. Additionally, the code also has features to help doctors capture errors in real-time which will be used for reinforcement learning later. By leveraging these technologies, the project aims to provide a robust, accurate, and cost-efficient solution for tracking surgical gauze.

## 4 RESULTS

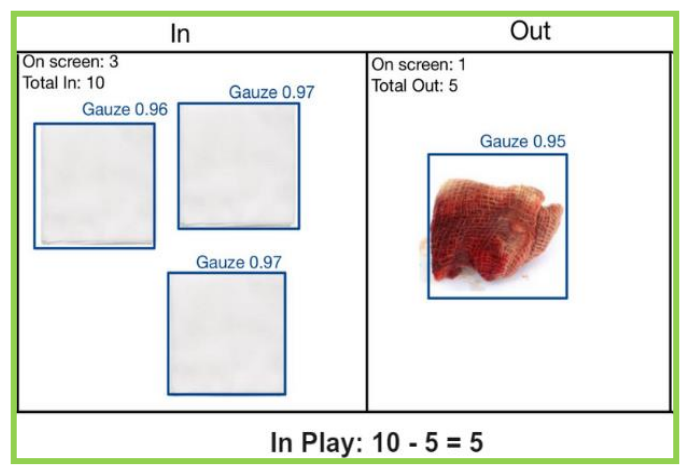

Figure 6: Output Screen UI

Currently the code runs in real time and is connected to a TV screen to display the output. As seen in Figure 6, it shows the output screen (User Interface), which is divided into half- one each showcasing the "In" and "Out" tray. The algorithm captures the gauzes being placed on the trays keeps a count of them. "On Screen" tells the number of gauzes currently as seen on the trays. "Total In" represents the total number of gauzes which have already been put in the "In" tray since the beginning of the operation and "Total Out" represents the total number of gauzes which have been placed on the "Out" tray after being utilized for operational purposes.

"In Play" = "Total In - Total Out", which represents the number of gauzes which are currently inside the patient's body/ being used in the operation.

We have also created a "Traffic Lighting System" wherein the green border indicates opposite actions wherein the user can go ahead and put a gauze in the trays or take a gauze out from the trays. A yellow border signifies that a hand has been detected by the system and a red border, which usually appears only for about 50 milliseconds signifies that the system is updating the counts in the backend. This has been created for the User experience of the doctors and nurses to be seamless and more intuitive.

The aim is that after the operation concludes "Total In" and "Total Out" have the same value while "In Play=0", representing that there are no gauzes left behind in the patient's body.

## 5 DISCUSSION

Initially, the system utilized two distinct models for detecting gauzes and hands. However, we have now developed an integrated model that can detect both gauzes and hands, with an eightfold increase in our dataset size. This enhancement has significantly improved the model's precision and FPS rate. Our current setup involves connecting cameras to a laptop running our Python code. In the future, we aim to upgrade the 2 Tray system by incorporating a TV screen with the model running in the backend on a disc, eliminating the need for an external laptop. After testing with an additional hospitals in Singapore to gather more feedback and refine our product, we plan to commercialize it.

Initially, the AI dual-tray system offers highly accurate surgical gauze counts at a fraction of the cost, with adjustable software expenses. The primary cost factors are the hardware components, such as cameras and trays. We are confident that the efficiency of our system will motivate hospitals to adopt this AI gauze counting solution, eliminating the errors associated with manual counting and the high costs of RFID systems. As a result, operations can be conducted with greater accuracy at a lower cost. Additionally, reducing the incidence of Gossypiboma will enhance patient safety in the operating room, thereby boosting the hospital's reliability and reputation for superior patient care.

A thorough examination of medical databases reveals that the financial impact of Gossypiboma on healthcare facilities can amount to an average of $1.46 million, primarily due to legal settlements from patient litigations[7]. For example, headlines such as "Kentucky Woman Awarded $10.5 Million After Surgeons Left a Sponge Inside Her" underscore the substantial monetary and reputational harm hospitals incur from RSI incidents[8]. The introduction of this AI system has the potential to significantly decrease such incidents, thereby reducing the necessity for expensive legal disputes and their associated costs.

Nonetheless, the current feedback from nurses have raised concerns that even though the machine is counting accurately, they need to separate the gauzes and put them on the tray one by one on the "In" tray before the gauze enters the patients body. The also wished that the system could count multiple stacked gauzes kept together. Currently, it can detect gauzes correctly if there is an overlap of around 50% between gauzes and in the future iterations of this project we would try to include more sophisticated models which will be correctly able to detect stacked gauzes too.

In summary, this system offers advantages as an economical and dependable method for everyone involved in surgical procedures, including patients, medical professionals, and healthcare facilities.

## 6 ACKNOWLEDGEMENTS

I am grateful to my project supervisor, Dr. Cai Yiyu, for providing me with the opportunity to work on this project and his continuous support throughout. This gave me a unique opportunity to learn and be a part of the AI advancements in the healthcare industry. Additionally, I would like to thank Dr Huang Li Hui for coordinating this project and helping us will any necessary information always. I would also like to thank Dr. Tay from SGH hospitals for his commitment to this project and always giving us essential insights to make our system an efficient one. My PhD mentor, Azam Abu Bakr helped me to understand the intricacies of the code and FYP student Victor Raj guided me and made significant contributions to this project. I would like to express my gratitude to SGH and NTU to provide grants for this research.